\documentclass[12pt]{article}
\pdfoutput=1

\usepackage{graphicx}
\usepackage[utf8]{inputenc}
\usepackage{amsmath}
\usepackage{fancyhdr}
\usepackage{geometry}
\usepackage{centernot}
\usepackage{tabularx}
\usepackage{makecell}
\usepackage{placeins}
\usepackage[labelfont=bf,labelsep=period]{caption}

\usepackage[dvipsnames]{xcolor}
\usepackage[linguistics,edges]{forest}
\usetikzlibrary{shadows,arrows.meta}
\usepackage[frozencache,cachedir=./_minted-output]{minted}
\usepackage{etoolbox}
\patchcmd{\quote}{\rightmargin}{\leftmargin 4em \rightmargin}{}{}

\edef\restoreparindent{\parindent=\the\parindent\relax}
\usepackage{parskip}
\restoreparindent

\newcommand{\ite}[3]{\ensuremath{\left(#1 \: ? \: #2 : #3\right)}} 
\newcommand{\cond}[3]{\ensuremath{[#3]\left(#1\! \mid\! #2\right)}}
\newcommand{\expect}{E}

\begin{document}
\title{\vspace{-3em}Tyche: A library for probabilistic reasoning and belief modelling in Python}
\author{\vspace{0.4em}Padraig X. Lamont \\ \small{School of Engineering, The University of Western Australia}}
\date{}

\maketitle

\begin{abstract}
    
This paper presents Tyche, a Python library to facilitate probabilistic reasoning in uncertain worlds through the construction, querying, and learning of belief models. Tyche uses aleatoric description logic (ADL), which provides computational advantages in its evaluation over other description logics. Tyche belief models can be succinctly created by defining classes of individuals, the probabilistic beliefs about them (concepts), and the probabilistic relationships between them (roles). We also introduce a method of observation propagation to facilitate learning from complex ADL observations. A demonstration of Tyche to predict the author of anonymised messages, and to extract author writing tendencies from anonymised messages, is provided. Tyche has the potential to assist in the development of expert systems, knowledge extraction systems, and agents to play games with incomplete and probabilistic information.

\vspace{0.8em}

\end{abstract}

\section{Introduction}
Many of the major advances in artificial intelligence in the last decade have embraced the use of data that does not follow hard rules \cite{LINN_NEURAL_LOGIC_REASONING,IMAGENET_2012,GPT3,ALPHA_FOLD}. Systems such as the use of deep CNNs for ImageNet classification, AlphaFold, and GPT-3 have all demonstrated incredible performance over hard rules-based systems for handling data containing a high degree of uncertainty \cite{IMAGENET_2012,ALPHA_FOLD,GPT3}. However, many of our existing ontological knowledge base systems are designed only to hold facts about our world, with no allowance for uncertainty \cite{PROTEGE,LOGIC_AND_ONTOLOGY,OWL}. Tyche aims to facilitate probabilistic reasoning about uncertain information through the use of aleatoric description logic, and through the creation of ontological knowledge bases with probabilistic beliefs (i.e., belief models).

Aleatoric description logic (ADL) provides a novel probabilistic description logic to represent and reason about uncertain information \cite{ADL}. ADL models belief by rolls of dice, with queries about the world believed to be true only a percentage of the time. For example, we may believe that there is a 30\% chance that it will rain on any given day.  This interpretation of probability distinguishes ADL from other probabilistic logics, which use probability to represent a degree of belief about the truth of facts \cite{LOGIC_AND_PROBABILITY,PROBLOG}, the state of the world \cite{BLOG}, or, in the case of fuzzy logics, the degree of inclusion to a set \cite{FUZZY_LOGIC}. Furthermore, this independence in the sampling of beliefs in ADL leads to a simple recursive interpretation of its semantics, which gives Tyche computational advantages over other probabilistic logic languages that use logical solvers \cite{PROBLOG,BLOG}.

Aleatoric description logic is evaluated in Tyche using belief models to provide the probabilistic beliefs about individuals (termed concepts), and the probabilistic relationships between individuals (termed roles). Belief models are created within Tyche through the definition and instantiation of classes of individuals. The classes of individuals define the concepts and roles that are available for a type of individual. The value of these concepts and roles are supplied by the individuals themselves (i.e., the instances of the individual classes). These values may be supplied by fields in a class, or generated dynamically by the methods of the class. For example, the probabilistic beliefs about an individual may be calculated from their non-probabilistic properties using other tools such as continuous probability distributions, expert rules, or neural networks. This grants a large degree of flexibility when defining belief models using Tyche. This was a key goal in Tyche's implementation, to facilitate interoperability between Tyche and other systems. The goals of flexibility and interoperability when developing Tyche also guided the choice to implement Tyche in Python, due to its vast ecosystem of available tools for scientific computing \cite{PYTHON_ECOSYSTEM}.

\section{Aleatoric Description Logic}
Aleatoric description logic (ADL) provides a generalisation of standard description logics to extend true/false information to the closed interval [0, 1], with the extended values representing probabilities of truth \cite{ADL}. The base syntax of ADL consists of the constant $always$ ($\top$), the constant $never$ ($\bot$), atomic concepts (named variables), the ternary operator ($(\alpha \: ? \: \beta : \gamma)$), and the marginalisation operator ($[\rho](\alpha|\beta)$). The function of these syntax elements of ADL are described in Table \ref{tab:adl_syntax}. These base syntax elements provide a versatile basis to construct common syntactical elements from other description logics, as shown in Table \ref{tab:adl_abbreviations}.

\bgroup
\setlength{\tabcolsep}{0.5em}
\def\arraystretch{1.1}%
\begin{table}[ht!]
\begin{center}
    \caption{Descriptions of the base syntax elements of aleatoric description logic.}
    \label{tab:adl_syntax}
    
    \begin{tabularx}{\textwidth}{ |c|X|c| } 
    
    \hline
    \textbf{Name}
    & \textbf{Description}
    & \textbf{ADL} \\ 
    
    \hline
    Always
    & The constant true, or 100\%.
    & $\top$ \\ 
    
    \hline
    Never
    & The constant false, or 0\%.
    & $\bot$ \\ 
    
    \hline
    Atomic Concepts
    & A sampling of a probabilistic belief about the world.
    & Named Variables \\ 
    
    \hline
    Ternary Operator
    & A ternary operator that evaluates to $\beta$ if $\alpha$ is sampled as true, or else evaluates to $\gamma$.
    & $(\alpha \: ? \: \beta : \gamma)$ \\ 
    
    \hline
    Marginalisation Operator
    & The marginalisation operator that evaluates to the chance that $\alpha$ is true, given that an individual in the role $\rho$ was encountered for which $\beta$ was true.
    & $[\rho](\alpha|\beta)$ \\
    
    \hline
    \end{tabularx}
\end{center}
\end{table}
\egroup

\bgroup
\setlength{\tabcolsep}{0.5em}
\def\arraystretch{1.1}%
\begin{table}[!ht]
\begin{center}
    \caption{List of common abbreviations used in description logics, and their equivalent ADL sentences.}
    \label{tab:adl_abbreviations}
    
    \begin{tabular}{ |c|c|c| } 
    
    \hline
    \textbf{Name}
    & \textbf{Abbreviation}
    & \textbf{Equivalent ADL Sentence} \\ 
    \hline
    Conjunction & $\alpha\land\beta$ & $\ite{\alpha}{\beta}{\bot}$ \\
    \hline
    Disjunction & $\alpha\lor\beta$ & $\ite{\alpha}{\top}{\beta}$ \\
    \hline
    Complement & $\lnot\alpha$ & $\ite{\alpha}{\bot}{\top}$ \\
    \hline
    Implication & $\alpha\Rightarrow\beta$ & $\ite{\alpha}{\beta}{\top}$ \\
    \hline
    Expectation & $\expect_\rho\alpha $&$ \cond{\alpha}{\top}{\rho}$ \\
    \hline
    Existential & $\exists\rho.\alpha $&$ \lnot\cond{\bot}{\alpha}{\rho}$ \\  
    \hline
    \end{tabular}
\end{center}
\end{table}
\egroup

An important feature of ADL is that each occurrence of a concept or role within an aleatoric description logic sentence is treated as an independent sampling of the value of that concept or role \cite{ADL}. \textit{The probabilities of concepts and roles do not represent the chance that their underlying value is true, but instead the chance that they will be true when sampled.} For example, if the probability of $\alpha$ is not 0 or 100\%, then the probability of $(\lnot \alpha) \land \alpha$ is not 0\%. This limits the simplification that can be performed on ADL sentences, but also allows ADL sentences to be evaluated recursively, without a complicated solver. This is an important feature of ADL, as this allows Tyche to model different types of problems from other probabilistic logics.

\section{Quantification of Belief: Concepts and Roles}
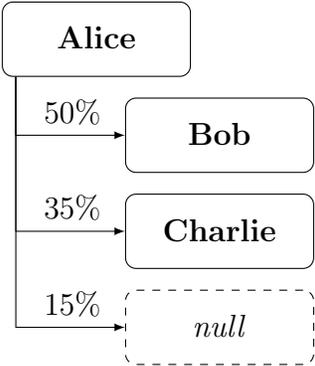
\begin{figure}[ht!]
\begin{center}
\begin{forest}
  arrow label/.style={
    edge label={node [midway, anchor=south west] {#1}},
  },
  for tree={
    minimum width=2.5cm,
    edge={->,>=latex},
    rounded corners,
    draw,
    inner sep=5,
    if level=0{
      child anchor=north,
      for tree={
        grow'=0,
        folder,
        l sep'+=30pt,
      },
    }{},
  },
  before typesetting nodes={
    for tree={
      content/.wrap value=\strut {#1},
    },
  },
  [{\textbf{Alice}}
    [{\textbf{Bob}}, tier=b, arrow label={\: 50\% \:}]
    [{\textbf{Charlie}}, tier=b, arrow label={\: 35\% \:}]
    [{\textit{null}}, dashed, tier=b, arrow label={\: 15\% \:}]
  ]
\end{forest}
\end{center}
\caption{An example role relating Alice to other individuals, or the null individual.}
\label{fig:role_tree}
\end{figure}

Tyche supports reasoning about probabilistic beliefs about both the properties of individuals (termed concepts), and the relationships between individuals (termed roles). The value of concepts are provided as floating-point probability values in the range [0, 1]. The value of roles are provided as mutually exclusive probability distributions of potentially related individuals. Roles may also include the null-individual to represent ``no relation''. An example value of a role between individuals is shown in Fig. \ref{fig:role_tree}.

The value of concepts and roles may be dynamically calculated by belief models, or they may be supplied as constants. For example, the probability of an individual being tall may be dynamically calculated from their expected height distribution and a height cut-off (see Fig. \ref{code:person}). This aims to facilitate the combined use of Tyche with other methods to calculate probabilities, such as continuous probability distributions, expert rules, or neural networks.

\section{Representation of Aleatoric Description Logic}
\begin{figure}[ht!]
\begin{center}
\begin{forest}
  for tree={
    if n children=0{
      font=\itshape,
      tier=terminal,
    }{},
    minimum width=2.5cm,
    fit=band,
    l'=0pt,
    edge={->,>=latex},
    rounded corners,
    draw,
    inner sep=5,
  },
  forked edges,
  [{\textbf{Expectation} \\ $[\rho]((\alpha \: ? \: \beta : \bot) | \top)$}
    [{\textbf{Role} \\ $\rho$}, tier=b]
    [{\textbf{Conditional} \\ $\alpha \: ? \: \beta : \bot$}, tier=b
      [\textbf{Concept} \\ $\alpha$, tier=c]
      [\textbf{Concept} \\ $\beta$, tier=c]
      [\textbf{Constant} \\ $\bot$, tier=c]
    ]
    [{\textbf{Constant} \\ $\top$}, tier=b]]
  ]
\end{forest}
\end{center}
\caption{Tree representation of an example ADL sentence. The node class names are shown in bold, and below them are the sentences represented by the nodes.}
\label{fig:adl_tree}
\end{figure}
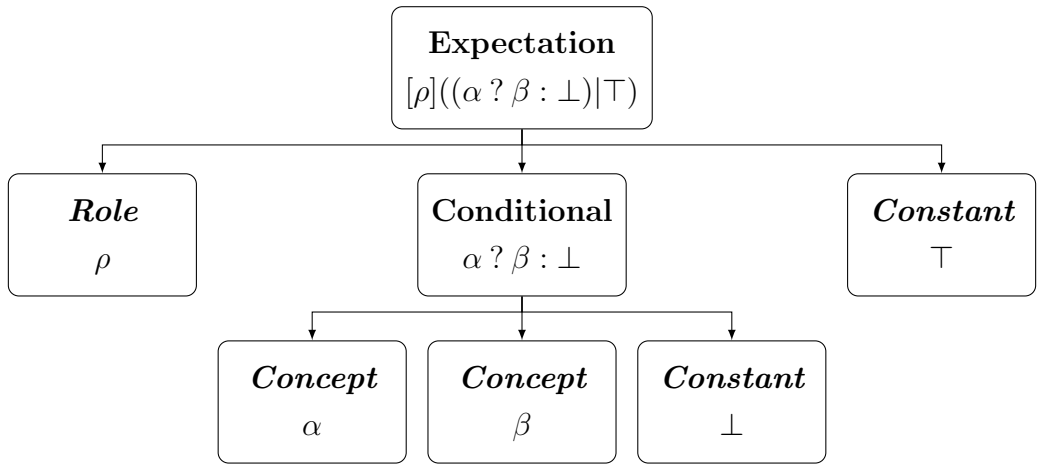

Aleatoric description logic (ADL) sentences are represented as trees within Tyche. Each element within an ADL sentence is represented as a node in the tree, with the node's constituent subsentences represented as child nodes. The tree structure of an example ADL sentence, $[\rho]((\alpha ? \beta : \bot) | \top)$, is illustrated in Fig. \ref{fig:adl_tree}. This example sentence represents the expectation that an individual selected from role $\rho$ will have true values sampled for the concepts $\alpha$ and $\beta$. For example, this could represent the expectation that a friend you ran into was happy and relaxed. This example sentence may be created using Tyche using the following code, $Expectation(``\rho\text{''}, \text{\:} Concept(``\alpha\text{''}) \text{\: \& \:} Concept(``\beta\text{''}))$.

\vspace{0.4cm}

\newpage
\subsection{ADL Constructs}
Tyche provides several aleatoric description logic (ADL) constructs for use in the creation of ADL sentences. Each construct is represented by their own node class in the language module of Tyche. All the node classes except Role are sub-classes of the ADLNode base class, which is used to represent ADL sentence trees that evaluate to probabilities. Roles are treated specially, as they evaluate to a mutually exclusive probability distribution of individuals, and not probabilities. The node types, their descriptions, and their ADL representations are shown in Table \ref{tab:adl_node_types}.

\bgroup
\setlength{\tabcolsep}{0.5em}
\def\arraystretch{1.1}%
\begin{table}[ht!]
\begin{center}
    \caption{Descriptions of the node types provided by Tyche, along with their ADL representation.}
    \label{tab:adl_node_types}
    
    \begin{tabularx}{\textwidth}{ |c|X|c| } 
    
    \hline
    \textbf{Node Class}
    & \textbf{Description}
    & \textbf{ADL} \\ 
    
    \hline
    Concept
    & A sampling of the value of a concept from a belief model.
    & Named Variables \\ 
    
    \hline
    Role
    & A sampling of the value of a role from a belief model.
    & Named Variables \\ 
    
    \hline
    Constant
    & Fixed probabilities, such as a coin flip (50\%), always (100\%), or never (0\%).
    & \makecell{$\top$ (always), \\ $\bot$ (never)} \\ 
    
    \hline
    Conditional
    & A ternary operator that evaluates to $\beta$ if $\alpha$ is true, or else evaluates to $\gamma$. These operators can be used to construct several common logical operators such as AND, OR, or NOT.
    & $(\alpha \: ? \: \beta : \gamma)$ \\ 
    
    \hline
    Expectation
    & A marginalisation operator that evaluates to the chance that $\alpha$ is true, given that an individual in the role $\rho$ was encountered for which $\beta$ was true. This ignores the presence of the null-individual, and is vacuously true if the role contains no non-null individuals.
    & $[\rho](\alpha|\beta)$ \\ 
    
    \hline
    Exists
    & An operator that evaluates to the chance that a role has a related individual (i.e., a relation that is not the null-individual).
    & N/A \\
    
    \hline
    \end{tabularx}
\end{center}
\end{table}
\egroup

\section{Evaluation of Aleatoric Description Logic}
The probability of truth of aleatoric description logic (ADL) sentences may be evaluated about an individual in a belief model. The result represents the chance that the sentence will be true when sampled about that individual (that is, the limit of a Monte Carlo estimation). Due to the nature of aleatoric description logic, each subsentence within an ADL sentence may be evaluated independently \cite{ADL}. This allows Tyche to evaluate ADL sentences recursively, without a logical solver. The evaluation of ADL sentences in Tyche follows the semantics of Modal Aleatoric Calculus \cite{MAC}.

\subsection{Evaluation Contexts}
The evaluation of ADL sentences requires a context to provide the value of concepts and roles. Individuals from Tyche belief models are intended to be used as these contexts for the evaluation of ADL queries. However, custom contexts may also be defined by subclassing the TycheContext class. The evaluation context may change if a marginalisation operator is used, as the subsentences of the expectation will be evaluated with each of the related individuals as the context, instead of the current context.

\subsection{Evaluation Procedures}
ADL sentences can be evaluated by providing them to the $eval$ method of an evaluation context. The context may then invoke the ADL nodes' own evaluation procedures. This redirection gives the contexts the power to change the method of evaluation for ADL sentences. The evaluation procedures for each ADL construct provided by Tyche is shown in Table \ref{tab:evaluation_procedures}.

\bgroup
\setlength{\tabcolsep}{0.4em}
\def\arraystretch{1.1}%
\begin{table}[ht!]
\begin{center}
    \caption{The evaluation procedure of each ADL node type provided by Tyche.}
    \label{tab:evaluation_procedures}
    \begin{tabularx}{\textwidth}{ |c|X| } 
    
    \hline
    \textbf{Node Class}
    & \textbf{Evaluation Procedure} \\ 
    
    \hline
    Concept
    & The value of concepts are accessed from the current context. \\
    
    \hline
    Role
    & The value of roles are accessed from the current context. \\
    
    \hline
    Constant
    & The fixed probability of the constant is returned. \\
    
    \hline
    Conditional
    & 
        For the ternary $(\alpha \: ? \: \beta : \gamma)$, the probability of the result is calculated following modal aleatoric calculus, \cite{MAC}
        \newline
        $$P(\alpha \: ? \: \beta : \gamma) = P(\alpha) \cdot P(\beta)+(1-P(\alpha)) \cdot P(\gamma)$$
    \\
    
    \hline
    Expectation
    &
        The marginalisation operator, $[\rho](\alpha|\beta)$, evaluates the role $\rho$ in the current context, and $\alpha$ and $\beta$ in related contexts. The final expected probability is calculated following,
        \newline
        $$P([\rho](\alpha|\beta))=\frac{\sum_{x \in \rho} P(x \: \text{selected from} \: \rho) \cdot P_x(\alpha \land \beta)}{\sum_{x \in \rho} P(x \: \text{selected from} \: \rho) \cdot P_x(\beta)}$$
        \newline
        However, if the sum in the denominator is zero, (i.e., there are no possible related individuals), then the marginalisation operator is vacuously true and, as such, evaluates to 100\%.
     \\
    
    \hline
    Exists
    &
        The role $\rho$ is evaluated in the current context, and the value of the exists operator is evaluated following,
        \newline
        $$P(Exists(\rho))=1 - P(\text{null selected from} \: \rho)$$
    \\
    
    \hline
    \end{tabularx}
\end{center}
\end{table}
\egroup

\newpage
\section{Constructing Belief Models}
Tyche supports the creation of belief models as ontological knowledge bases of individuals, the probabilistic beliefs about them (concepts), and the probabilistic relationships between them (roles). Python classes may be defined to represent the possible types of individuals by subclassing Individual, and their fields and methods may be type-annotated or decorated, respectively, to register them as providing the value of concepts or roles. This allows individuals to be flexibly defined with support for polymorphism of the individual classes. The code used to define an example individual called Person is shown in Figure \ref{code:person}.

\begin{figure}[ht!]
\begin{quote}
\vspace{0.3cm}
\begin{minted}{python3}
from tyche.distributions import *
from tyche.individuals import *
from tyche.language import *

class Person(Individual):
    positive: TycheConceptField
    conversed_with: TycheRoleField

    def __init__(self, positive: float, height_cm: NormalDist):
        super().__init__()
        self.positive = positive
        self.conversed_with = ExclusiveRoleDist()
        self.height_cm = height_cm

    @concept(symbol='tall')
    def is_tall(self):
        return self.height_cm > 180
\end{minted}
\vspace{0.2cm}
\end{quote}
\caption{Example definition of a type of individual called Person that has a concept ``positive'' provided by a field, a concept ``tall'' provided by a method, and a role ``conversed\_with'' provided by a field and initialised as empty.}
\label{code:person}
\end{figure}

\subsection{Registering fields as concepts or roles}
The fields of an Individual subclass can be marked as providing the value of a concept or role by type-annotating them with the ``TycheConceptField'' type, or the ``TycheRoleField'' type, respectively. This will inform Tyche to access the value of these fields to use as the value for the concepts and roles in your ADL queries. The names of the fields will be used as the symbols for the concepts and roles within your ADL queries. The ``is\_positive'' field and the ``conversed\_with'' field in Fig. \ref{code:person} are registered as a concept and role, respectively, using this method.

\subsection{Registering methods as concepts or roles}
The methods of an Individual subclass can be marked as providing the value of a concept or role by decorating them with the ``@concept()'' decorator, or the ``@role()'' decorator, respectively. These methods cannot take any arguments and should return the value of the concept or role. The name of the method will be used as the symbol of the concepts and roles by default, although a symbol can also be explicitly provided to the decorators through their symbol parameter. The ``is\_tall'' method in Figure \ref{code:person} is registered as the concept ``tall'' using this method.

\section{Learning through Observation}
Tyche provides mechanisms to learn the value of concepts and roles from observations of their values. The observations take the form of ADL sentences, and can be supplied to individuals using their ``observe'' method. Tyche will use Bayes' rule to determine how each term in the ADL sentence has influenced the truth of the entire sentence. The values of the concepts and roles used within the sentence may then be learnt using one of the learning strategies supplied by Tyche. These learning strategies must be provided through a method decorator. The code used to define an example individual named Student with learning registered for its ``good\_grades'' concept is shown in Fig. \ref{code:student}.

\begin{figure}[ht!]
\begin{quote}
\vspace{0.3cm}
\begin{minted}{python3}
class Student(Person):
    def __init__(self, good_grades: float):
        super().__init__(0.33, NormalDist(175, 6.5))
        self._good_grades = good_grades

    @concept()
    def good_grades(self):
        return self._good_grades

    @good_grades.learning_func(DirectConceptLearningStrategy())
    def set_good_grades(self, good_grades: float):
        self._good_grades = good_grades
\end{minted}
\vspace{0.2cm}
\end{quote}
\caption{Example definition of a type of individual called Student with a concept ``good\_grades'' that may be learnt using the direct concept learning strategy. The Student type also inherits the concepts and roles of the Person type from Fig. \ref{code:person}.}
\label{code:student}
\end{figure}

\subsection{Calculation of the influence of terms in ADL observations}
The influence of each term in ADL observations is calculated recursively, through the propagation of two parameters: likelihood and learning rate. The likelihood parameter represents the chance that the current term in the observation is true, and the learning rate quantifies the percentage impact of the current term on the truth of the entire sentence. These two parameters allow the influence of each term in an ADL observation to be calculated through a simple algorithm that recurses through the ADL observation tree. This is important, as it means that learning strategies only need to deal with simple observations that are directly related to the concept or role being learnt (e.g., $is\_sunny$, or $[friend](is\_tired)$). The learning strategies do not need to deal with the interactions between terms in complex ADL observations. The independence of each term in ADL sentences is crucial for this method of influence propagation to work.

\subsubsection{Calculation of the Likelihood Parameter}
The likelihood of each term in an ADL observation can be calculated using a method built upon the application of Bayes' rule. Initially, the likelihood parameter for an observation is given a value of $1$. This represents that the observation itself is true. The likelihood of each child node of the original observation ($obs$) can then be calculated using Bayes' rule as in Equation \ref{eqn:bayes_rule}.

\begin{align}
\label{eqn:bayes_rule}
P(child|obs) &= \frac{P(obs|child) \cdot P(child)}{P(obs)}
\end{align}

This method allows the likelihood parameter of each term of the original observation to be calculated. However, this calculation relies on the truth of the observation (i.e., it requires $likelihood=1$). This requirement does not necessarily hold for each node in an ADL observation tree. Therefore, to propagate the likelihood through the entire observation tree, we must also account for the chance that the parent node is not true (i.e., $likelihood < 1$). We can do this by applying Bayes' rule to the event that either the parent node was true with chance $\alpha=likelihood$, or else the parent node was false \cite{BAYES_UNCERTAIN_EVIDENCE_SO}. This results in a weighted sum based upon the likelihood, $\alpha$, as in Equation \ref{eqn:uncertain_bayes_rule}.

\begin{align}
\label{eqn:uncertain_bayes_rule}
P(child|parent) &=
\alpha \cdot P(child \:|\: parent) + 
(1 - \alpha) \cdot P(child \:|\: \lnot parent)
\\[1.4em]
&= \alpha \cdot \frac{P(parent \:|\: child) \cdot P(child)}{P(parent)} \notag \\[0.8em]
&{\hspace{12pt}} + (1 - \alpha) \cdot \frac{(1 - P(parent \:|\: child)) \cdot P(child)}{1 - P(parent)}
\end{align}

The value of $P(child)$ and $P(parent)$ may be calculated directly from the current belief model. The value of $P(parent|child)$ may be calculated by replacing the child in the observation with $\top$ (i.e., a constant 100\%), and evaluating the resulting sentence in the current belief model. 

\subsubsection{Calculation of the Learning Rate Parameter}
The learning rate of each node in an ADL observation tree can be calculated using the learning rate of its parent, and the difference in truth of its parent if the term were true, and if it were false. Initially, the learning rate parameter for an observation is given a value of $1$. The resulting learning rate, $r$, of each child node may then be calculated following Equation \ref{eqn:learning_rate}.

\begin{align}
\label{eqn:learning_rate}
r_{child} &= r_{parent} \cdot abs(P(parent|child) - P(parent|\lnot child))
\end{align}

This value computes the influence that the child had on its parent, and multiples that by the parent's influence, to determine the influence of the child on the original observation. However, when an expectation over a role is reached, then this calculation can no longer be performed. Instead, the chance that each individual within the role was selected is used to modulate the learning rate that is propagated to each individual within the role. For an expectation $[\rho](a|b)$, the chance that each individual, $x$, was selected within the role, $\rho$, is given by Equation \ref{eqn:reverse_observation}.

\begin{align}
\label{eqn:reverse_observation}
&P(x \: \text{selected from} \: \rho \: \text{for} \: obs) = \notag \\[0.8em]
&{\hspace{24pt}} \frac{P(x \: \text{selected from} \: \rho) \cdot P_x(b) \cdot (\alpha \cdot P_x(a) + (1 - \alpha) \cdot (1 - P_x(a)))}{\sum_{y \in \rho} P(y \: \text{selected from} \: \rho) \cdot P_y(b) \cdot (\alpha \cdot P_y(a) + (1 - \alpha) \cdot (1 - P_y(a)))}
\end{align}

This chance can then be used as a multiplier with the learning rate of the expectation node, $\alpha$, to calculate the learning rate to propagate to each related individual.

\subsection{Design of learning strategies}
Learning strategies update the value of concepts and roles based upon observations of their use. A few basic learning strategies are provided by Tyche, although custom learning strategies for concepts and roles may also be created by subclassing ConceptLearningStrategy or RoleLearningStrategy, respectively. Learning strategies do not receive the entire observation, but instead only receive the sub-sentence of the observation that is relevant to them. Each inclusion of a Concept node in an observation will be passed to the concept's learning strategy, if one is registered. Similarly, each inclusion of an Expectation node will be passed to its role's learning strategy, if one is registered. The learning strategies may then use the provided sub-sentence and the propagated influence parameters to update the current context (e.g., to update the current individual). This structure provides a simple method for writing learning strategies that may be re-used for learning the value of concepts and roles from many different types of ADL observations.

\subsection{Available learning strategies}
Tyche currently provides two learning strategies for concepts, and two learning strategies for roles. The Bayes' rule learning strategy updates the distribution of roles by updating the probability of each individual in a role to the conditional probability of selecting that individual, given the observation. The direct concept learning strategy updates the values of concepts to the weighted sum of the current value of the concept and the likelihood parameter, weighted by the learning rate. The statistical learning strategies maintain a running mean of the concept likelihoods or role distribution weights, which is used as the new value of the concepts or roles.

\subsubsection{Bayes' Rule Learning Strategy}
The Bayes' rule learning strategy (class \small{$BayesRuleLearningStrategy$}) uses a modified version of Bayes' rule to determine the new probability of each individual within a role based upon observations of a marginalisation operator (i.e., $[\rho](a|b)$). The conditional probability of each individual within the role, $P(x \text{\, selected from \,} \rho \: | \:obs)$, is calculated using Equation \ref{eqn:uncertain_bayes_rule}, which supports observations with any $likelihood$ parameter. The conditional probabilities of the individuals are used to construct a new role distribution. The value of the role is then set to the weighted sum of the current value of the role and this new role distribution, weighted by the learning rate, $r$.

\subsubsection{Direct Concept Learning Strategy}
The direct concept learning strategy (\small{$DirectConceptLearningStrategy$}) calculates the new concept value as the weighted sum of the concept's current value and the likelihood parameter, weighted by the learning rate, $r$.

$$v_{new}=r \cdot likelihood + (1 - r) \cdot v_{old}$$

If the learning rate is $1$, this results in observations of concepts setting the value of those concepts to the likelihood parameter. For example, an observation of $is\_hungry$ would result in $P(is\_hungry)=100\%$. This is likely not desired in an aleatoric context, so a $learning\_rate$ parameter may also be supplied to the learning strategy itself, which will be multiplied by $r$ before it is applied.

\subsubsection{Statistical Learning Strategies}
The statistical learning strategies maintain a running mean of a concept's likelihood, or a role's weights. As observations are made, the updated running means are used as the new value of the concepts or roles. Each observation that is added is weighted by its learning rate parameter. In concept learning (class \small{$StatisticalConceptLearningStrategy$}), a running mean of the received likelihood parameters is maintained. In role learning (class \small{$StatisticalRoleLearningStrategy$}), a running mean of the role distribution weights after the application of Bayes' rule, as in the Bayes' rule learning strategy, is maintained. Decay rate parameters may also be provided to discount old observations as new observations are made. The $decay\_rate$ parameter is the ratio used to discount old observations after each new observation is made. The $decay\_rate\_for\_decay\_rate$ is used as a weight to increase the $decay\_rate$ parameter after each observation towards $decay\_rate=1$. These parameters can improve the convergence of Tyche belief models when learning.

\section{Demonstration: Anonymised Messages}
A simple demonstration of Tyche was developed to demonstrate its prediction and learning capabilities for handling anonymised messages where the author of each message is unknown, but the recipient is known. Tyche is demonstrated to provide accurate predictions of the author of these anonymised sets of messages based upon the writing tendencies of the potential authors, which are quantified for messages using three properties: $uses\_emoji$, $capitalises\_first\_word$, and $is\_positive$. However, in some systems, we may not know the writing tendencies of the authors for use in their prediction. Therefore, Tyche's learning functionality is also demonstrated to learn the writing tendencies of each author, without knowing the messages that each author sent. The only information that Tyche is provided for this learning is the properties of each message, and the assumption that each individual would not receive messages from themselves.

\subsection{Ground-truth belief model}
The writing tendencies of three authors are specified in Table \ref{tab:gt_belief_model}. These authors are used as the ground-truth belief model for this demonstration. A role between the authors, $conversed\_with$, is also specified that relates the authors to the other authors, based upon who they are more likely to have received messages from. In this demonstration, each author is related to the other two authors uniformly, with a chance of 50\% for each other author.

\bgroup
\setlength{\tabcolsep}{0.5em}
\def\arraystretch{1.1}%
\begin{table}[ht!]
\begin{center}
    \caption{Ground-truth belief model containing the true message writing tendencies of the three example authors used in the anonymised messages demonstration.}
    \label{tab:gt_belief_model}
    
    \begin{tabular}{|c|c|c|c|c|c|c|c|c|c|c|} 
    
    \hline
    {} 
    & \textbf{Capitalises First Word (\%)}
    & \textbf{Is Positive (\%)}
    & \textbf{Uses Emoji (\%)} \\ 
    
    \hline
    \textbf{Bob} & 40.0 & 15.0 & 90.0 \\
    \hline
    \textbf{Alice} & 80.0 & 40.0 & 10.0 \\
    \hline
    \textbf{Jeff} & 50.0 & 50.0 & 50.0 \\
    
    \hline
    \end{tabular}
\end{center}
\end{table}
\egroup

\subsection{Predicting the author of sets of messages}
\begin{figure}[ht!]
    \begin{center}
        \scalebox{0.9}{
            \input{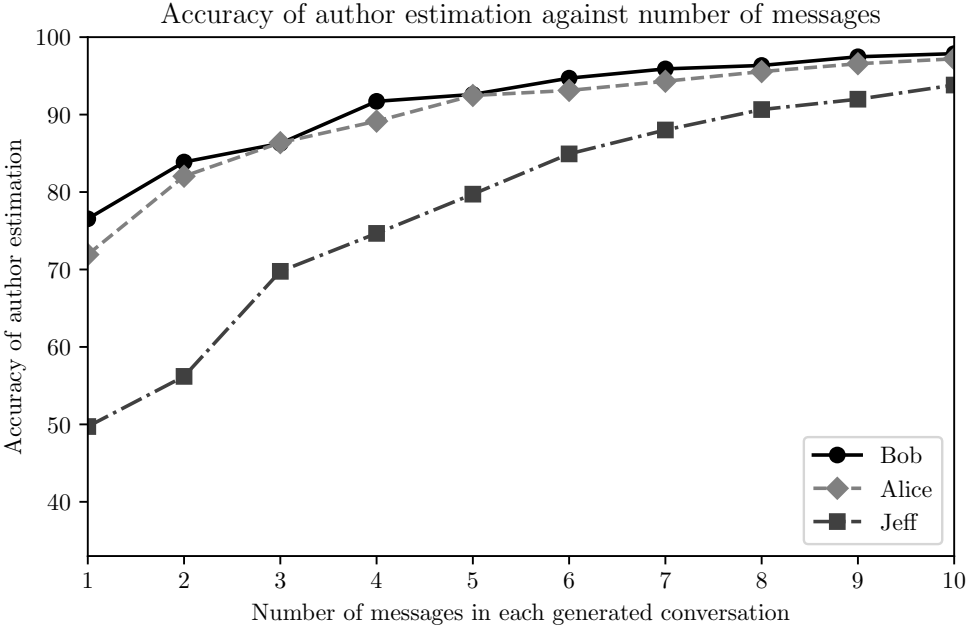}
        }
    \end{center}
    \caption{The accuracy of author estimation as the number of messages in each set of messages is increased, for the three individuals in the anonymised messages example.}
    \label{fig:author_prediction_accuracy}
\end{figure}

The author of sets of messages is predicted using Tyche by evaluating the chance that each author wrote the messages, based upon the ground-truth belief model. The author with the highest chance of having written the set of messages is selected as the predicted author. To evaluate Tyche's performance on this task, we used the ground-truth belief model to sample 100,000 random sets of messages from each author, with varying numbers of messages per set. The accuracy of Tyche at predicting the author of these random sets of messages is recorded in Table \ref{tab:author_pred}, and shown in Fig. \ref{fig:author_prediction_accuracy}.

\bgroup
\setlength{\tabcolsep}{0.5em}
\def\arraystretch{1.1}%
\begin{table}[ht!]
\begin{center}
    \caption{Author prediction accuracy (\%) for each individual in the anonymised messages example, with varying number of messages.}
    \label{tab:author_pred}
    
    \begin{tabular}{|c|c|c|c|c|c|c|c|c|c|c|} 
    
    \hline
    {} 
    & \multicolumn{10}{c|}{\textbf{Number of Messages}} \\ 
    
    \hline
    {} 
    & \textbf{1} & \textbf{2} & \textbf{3} & \textbf{4}
    & \textbf{5} & \textbf{6} & \textbf{7} & \textbf{8}
    & \textbf{9} & \textbf{10} \\ 
    
    \hline
    \textbf{Bob}
    & 76.5 & 83.9 & 86.3 & 91.7
    & 92.6 & 94.7 & 95.9 & 96.3 
    & 97.5 & 97.9 \\
    
    \hline
    \textbf{Alice}
    & 71.9 & 82.0 & 86.4 & 89.2
    & 92.5 & 93.1 & 94.3 & 95.6 
    & 96.6 & 97.2 \\
    
    \hline
    \textbf{Jeff}
    & 49.7 & 56.2 & 69.8 & 74.7
    & 79.7 & 84.9 & 88.0 & 90.7 
    & 92.0 & 93.6 \\
    
    \hline
    \end{tabular}
\end{center}
\end{table}
\egroup

\subsection{Learning the writing tendencies of authors}
The writing tendencies of the authors within a newly initialised belief model are learnt from observations of sets of messages received by each author. The initial belief model initialises all writing tendencies for all authors to 50\%. Each observation to be used in learning is then generated from the ground-truth belief model by first sampling the author that received the set of messages, then sampling who sent them the set of messages, and finally sampling the properties of the messages that they sent. These observations are then passed to the $observe$ method of the author that received the messages in the belief model being learnt. An example observation for a set of two messages received by an author in the belief model being learnt is shown in Equation \ref{eqn:example_observation}.

\begin{align}
\label{eqn:example_observation}
[conversed\_with](&(\lnot uses\_emoji \land \lnot capitalises\_first\_word \land is\_positive) \notag \\
&\land (\lnot uses\_emoji \land \lnot capitalises\_first\_word \land \lnot is\_positive))
\end{align}

To demonstrate the learning of belief models in Tyche, 100 trials were performed, with 500 total observations per trial, and 2 to 4 messages per observation. This corresponds to approximately 500 messages received per person, which is relatively small compared to the average number of messages received by people every year \cite{SMS_MESSAGES}. However, in this model, the 500 messages could only have been received from one of two people. At the beginning of each trial, the belief model is reset such that all author's writing tendencies are again set to 50\%. The number of messages in each observation was sampled from a uniform random distribution. The generated observations of messages were passed to the $observe$ method of the author that received the messages. The statistical concept learning strategy was used to learn all three writing tendencies for each author, with hyper-parameters $decay\_rate=0.95$ and $decay\_rate\_for\_decay\_rate=0.95$. The mean and standard deviation of the final writing tendencies learnt for each author at the end of each trial are shown in Table \ref{tab:learnt_belief_model}.

\bgroup
\setlength{\tabcolsep}{0.5em}
\def\arraystretch{1.1}%
\begin{table}[ht!]
\begin{center}
    \caption{The mean and standard deviations of the writing tendencies (\%) of the authors in the learnt belief models after 500 observations. The ground-truth writing tendencies are shown in brackets.}
    \label{tab:learnt_belief_model}
    
    \begin{tabular}{|c|c|c|c|c|c|c|c|c|c|c|} 
    
    \hline
    {} 
    & \textbf{Capitalises First Word (\%)}
    & \textbf{Is Positive (\%)}
    & \textbf{Uses Emoji (\%)} \\ 
    
    \hline
    \textbf{Bob}
    & 41.1 ± 4.6 \: \small{(40)}
    & 16.5 ± 3.3 \: \small{(15)}
    & 87.3 ± 3.1 \: \small{(90)} \\
    \hline
    \textbf{Alice}
    & 77.9 ± 3.8 \: \small{(80)}
    & 40.1 ± 4.4 \: \small{(40)}
    & 13.7 ± 3.9 \: \small{(10)} \\
    \hline
    \textbf{Jeff}
    & 50.4 ± 5.0 \: \small{(50)}
    & 46.6 ± 5.5 \: \small{(50)}
    & 50.3 ± 7.0 \: \small{(50)} \\
    
    \hline
    \end{tabular}
\end{center}
\end{table}
\egroup

The writing tendencies learnt for each author are quite similar to the ground-truth belief model from Table \ref{tab:gt_belief_model}. The error in the mean of the learnt writing tendencies is due to the number of observations being too low for Tyche to converge more closely to the ground-truth belief model. With more observations, the model is able to converge even more closely to the ground-truth belief model. This result demonstrates Tyche's efficacy to learn from indirect observations about the world. Therefore, we hope that Tyche may be used in similar ways to assist in the learning of unclear probabilistic properties within complex datasets.

\section{Related Work}
Tyche is tangentially related to work on other probabilistic logics such as ProbLog, Blog, and the use of fuzzy sets \cite{PROBLOG,BLOG}. ProbLog extends the Prolog programming language with the ability to specify the probability that each independent logical clause holds in a randomly sampled program \cite{PROBLOG}. ProbLog's assumption of the independence of the probability of clauses resembles ADL's assumption of the independence of the probabilities of concepts \cite{ADL}. However, the terms within clauses in ProbLog are not considered independent, whereas the terms in ADL sentences are considered independent. This leads to significant differences in the semantics of ProbLog and Tyche. ProbLog uses a solver to determine the probability that a query holds in a randomly sampled program, whereas Tyche recursively evaluates the probability of truth of a query, as if its terms were randomly sampled.

Blog provides a formal language to specify probability models, in a similar way to ProbLog \cite{BLOG,PROBLOG}. However, under Blog these probability models support worlds with unknown and unbounded numbers of objects, and identity uncertainty. This is a significant distinguishing factor from Tyche, which requires that all the individuals in the belief model, their concepts, and their roles, are defined. Additionally, Blog also requires the use of logical solvers to query its probability models, and does not consider terms in its sentences to be independent.

Fuzzy logic is used to represent degrees of truth, rather than a probability of truth as in ADL \cite{FUZZY_LOGIC}. For example, we do not usually consider objects as exclusively either ``warm'' or not. Instead, an object might be ``slightly warm'', ``moderately warm'', or ``very warm''. Fuzzy logic provides a mechanism to represent this, by representing knowledge as degrees of inclusion into a set. Therefore, the many-valued fuzzy logic serves a different purpose to ADL. However, despite this, they both share a similar functionality in Fuzzy Logic's use of t-norms \cite{FUZZY_LOGIC}. T-norms represent the belief that applying an uncertain hypothesis twice will lead to a different uncertainty than applying it only once. The t-norm operator of fuzzy logic shares a lot of functional similarities with ADL's operators, which provides evidence for the usefulness of repeated tests of a variable potentially yielding different results \cite{FUZZY_LOGIC,ADL}.

\section{Conclusion}
This paper has introduced Tyche, an open-source Python library to facilitate the use of aleatoric description logic (ADL) to reason about probabilistic belief models, and to support the creation and learning of them. We introduced Tyche's tree-based representation of ADL sentences, and its polymorphic representation of ontological knowledge bases of individuals and their relationships, for use as belief models. The method of recursive evaluation of the probability of truth of ADL queries using Tyche belief models was shown. We have also introduced an observation propagation algorithm to recursively calculate the influence of an ADL observation on each of its terms. These propagated influence parameters were shown to facilitate learning from complex ADL observations using simple learning strategies. Tyche's application to knowledge extraction was demonstrated through its use to predict the author of anonymised messages, and to learn the writing tendencies of authors without knowledge of the messages that they wrote. 

The source code of Tyche is available at https://github.com/TycheLibrary/Tyche.

\bibliographystyle{splncs04}
\bibliography{main}

\end{document}